\documentclass[runningheads]{llncs}
\usepackage[T1]{fontenc}
\usepackage{graphicx}
\usepackage{url}
\usepackage{color}
\usepackage{hyperref}
\usepackage[table,xcdraw]{xcolor}
\usepackage{amsmath}

\DeclareMathOperator{\argmin}{argmin}

\usepackage{amsfonts}
\usepackage{caption}
\usepackage{graphicx}
\usepackage{booktabs}
\usepackage{bbm}
\usepackage{wrapfig}
\usepackage{multirow}
\usepackage{capt-of}

\newcommand{\subgradp}{\phi}
\newcommand{\indexL}{l}
\newcommand{\R}{\mathbb{R}}
\newcommand{\valL}{L}
\newcommand{\norm}[1]{|\!| #1 |\!|}
\newcommand{\normd}[1]{\norm{#1}_{2}}

 {\begin{list}{}%
         {\setlength{\leftmargin}{#1}}%
         \item[]%
 }
 {\end{list}}

\begin{document}
\title{Multi-Modal Hypergraph Diffusion Network with Dual Prior for Alzheimer Classification }
%
%
\author
{
Angelica I. Aviles-Rivero\inst{1} \and
Christina Runkel \inst{1} \and
Nicolas Papadakis \inst{2}   \and \\
Zoe Kourtzi  \inst{3}   \and
Carola-Bibiane Schönlieb\inst{1}\thanks{the Alzheimer’s Disease Neuroimaging Initiative}
}


\authorrunning{Aviles-Rivero et al.}
%
\institute{
DAMTP, University of Cambridge, UK  \email{\{ai323,cr661,cbs31\}@cam.ac.uk}\and
IMB, Universite de Bordeaux, France \email{nicolas.papadakis@math.u-bordeaux.fr} \and
Department of Psychology, University of Cambridge, UK \email{zk240@cam.ac.uk}
}
\maketitle              
\begin{abstract}
The automatic early diagnosis of prodromal stages of Alzhei\-mer's disease is of great relevance for patient treatment to improve quality of life.  We address this problem as a multi-modal classification task. Multi-modal data provides richer and complementary information. However, existing techniques only consider lower order relations between the data and single/multi-modal imaging data. In this work, we introduce a novel semi-supervised hypergraph learning framework for Alzheimer’s disease diagnosis. Our framework allows for higher-order relations among multi-modal imaging and non-imaging data whilst requiring a tiny labelled set. Firstly, we introduce a dual embedding strategy  for constructing a robust hypergraph that preserves the data semantics. We achieve this by enforcing perturbation invariance at the image and graph levels using a contrastive based mechanism. Secondly, we present a dynamically adjusted hypergraph diffusion model, via a semi-explicit flow, to improve the predictive uncertainty. We demonstrate, through our experiments, that our framework is able to outperform current techniques for Alzheimer’s disease diagnosis.

\end{abstract}
%
%
%

\section{Introduction}
Alzheimer's disease (AD) is an irreversible, neurodegenerative disease impairing memory, language and cognition~\cite{de2016cellular}. It starts slowly but progressively worsens and causes approximately 60-80\% of all cases of dementia~\cite{alzheimers2022}. As there is no cure available yet, detecting the disease as early as possible, in prodromal stages, is crucial for slowing down its progression and for patient to improve quality of life.
The body of literature has shown the potentials of existing machine learning methods e.g.,~\cite{yang2021disentangled,polsterl2021scalable,shin2020gandalf}. However, there are two major limitations of existing techniques. Firstly, incorporating other relevant data types such as phenotypic data in combination with large-scale imaging data has shown to be beneficial e.g.~\cite{parisot2018disease}. However, existing approaches fail to  exploit the rich available imaging and non-imaging data. This is mainly due to the inherent problem of how to provide better and higher relations between multi-modal sources. Secondly, whilst vast amount of data is acquired every day at hospitals, labelling is expensive, time-consuming and prone to human bias. Therefore, to develop models that rely on extreme minimal supervision is of a great interest in the medical domain.

To address the aforementioned problems, hypergraph learning has been  explored -- as it allows going beyond pair-wise data relations. In particular, hypergraphs have already been explored for the task of AD diagnosis e.g.~\cite{pan2021characterization,zuo2021multimodal,shao2020hypergraph,shao2021hyper}. From the machine learning  perspective~\cite{berge1984hypergraphs},
several works have addressed the problem of hypergraph learning, by generalising the graph Laplacian to hypergraphs e.g.~\cite{zhou2006learning,hein2013total,saito2018hypergraph}. The principles from such techniques opened the door to extend the hypergraph Laplacian to graph neural networks (GNNs) e.g.~\cite{yadati2019hypergcn,feng2019hypergraph}.  However, the commonality of existing hypergraph techniques for AD diagnosis, and in general in the ML community, is that they consider the clique/star expansion~\cite{agarwal2006higher}, and in particular,  follow the  hypergraph normalised cut of that~\cite{zhou2006learning}.

To tackle the aforementioned challenges, we introduce a novel semi-supervised hypergraph learning framework for Alzheimer’s disease diagnosis. Our work follows a hybrid perspective, where we propose a new technique based on a dual embedding strategy and a diffusion model. To the best of our knowledge, this is the first work that explores invariance at the image and graph levels, and goes beyond the go-to technique of~\cite{zhou2006learning} by introducing a better hypergraph functional.

\textbf{Our contributions are as follows.} 1) We introduce a self-supervised dual multi-modal embedding strategy. The framework enforces invariance on two spaces-- the manifold that lies the imaging data and the space of the hypergraph structure. Our dual strategy provides better priors on the data distribution, and therefore, we construct a robust graph that offers high generalisation capabilities. 2) In contrast to existing techniques that follow~\cite{zhou2006learning}, we introduce a more robust diffusion-model. Our model is based on the Rayleigh quotient for hypegraph $p$-Laplacian and follows a semi-explicit flow.

\begin{figure}[t!]
\centering
\includegraphics[width=0.9\textwidth]{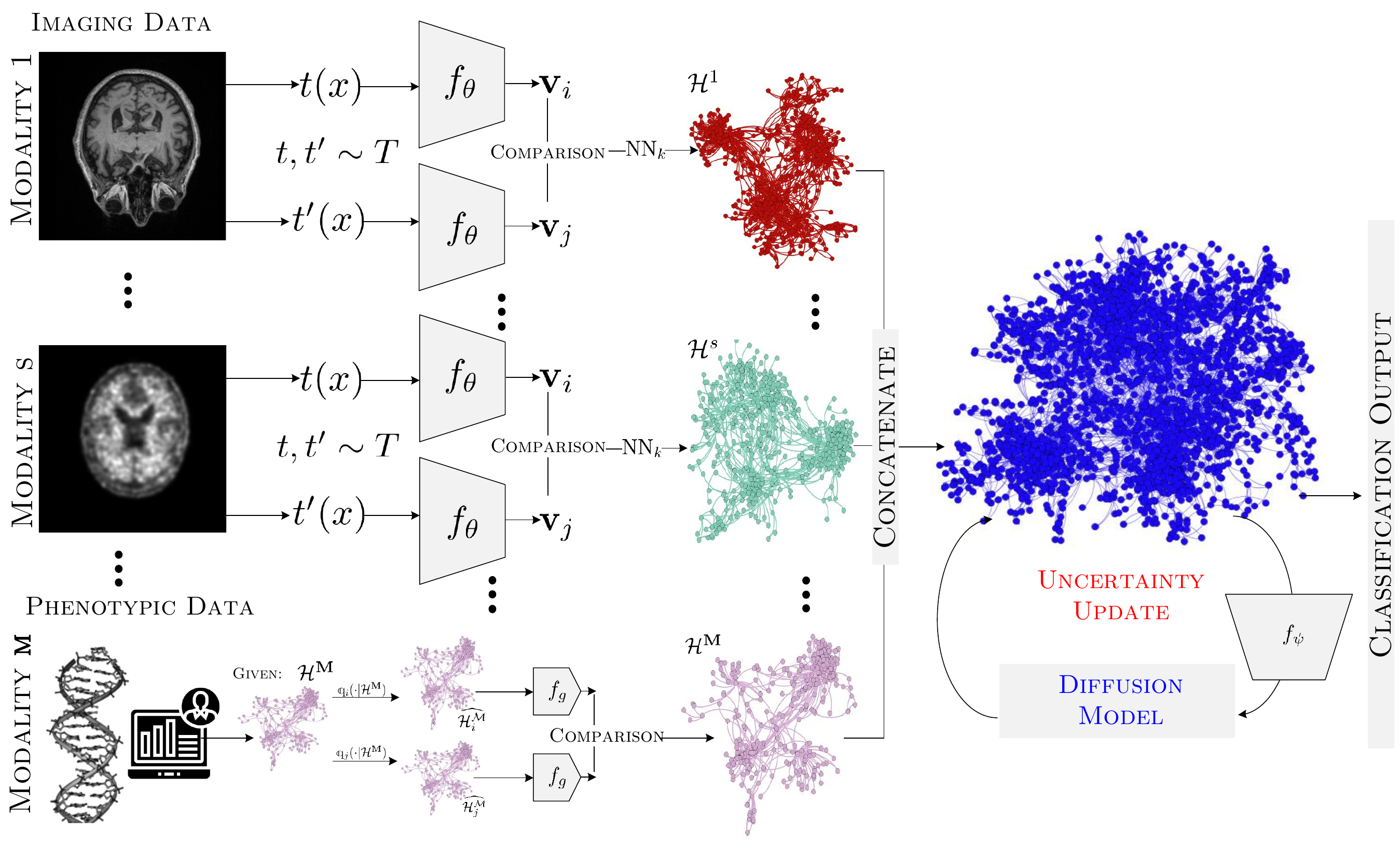}
\caption{Visual illustration of our framework. We first use a dual embeddings strategy at the image and hypergraph level. We then introduce a diffusion model for predicting early diagnosis of prodromal stages of Alzheimer’s disease.}
\label{fig::teaser}
\end{figure}

\section{Proposed Framework }
This section introduces our  semi-supervised hypergraph learning framework (see Fig.~\ref{fig::teaser}) highlighting two key parts: i) our dual embedding strategy to construct a robust graph, and ii) our dynamically adaptive hypergraph diffusion model.

\subsection{Hypergraph Embeddings \& Construction }
The first part of our framework addresses a major challenge in hypergraph learning -- that is,  how to extract meaningful embeddings, from the different given modalities, to construct a robust hypergraph. The most recent works for Alzheimer's disease diagnosis are based on extracting embeddings using sparse constraints or directly from a deep network. However, existing works only consider data level features to construct a graph or hypergraph and mainly for imaging data.
In contrast to existing works, we allow for higher-order relations between multi-modal imaging and non-imaging data.  Secondly, we enforce
a dual pertubation invariance strategy at both image and graph levels. With our dual
multi-modal embedding strategy, \textit{we seek to provide better priors on the data distribution such as our model is invariant to perturbations, and therefore, we can construct a robust graph that offers high generalisation capabilities.
}

We consider a hypergraph as a tuple $\mathcal{G} = (\mathcal{V},\mathcal{E}, w)$, where $\mathcal{V}= \{v_1, ..., v_n \}$, $|\mathcal{V}|=n$ is a set of nodes  and $\mathcal{E}= \{e_1, ..., e_m \}, \text{ } |\mathcal{E}|=m$ the hyperedges. Moreover, $w:\mathcal{E}\rightarrow \mathbb{R}_{>0}$ refers to the hyperedges weights, in which each hyperedge $e$ is associated to a subset of nodes. The associated incidence matrix is given by
$\mathcal{H}_{i,e}=  \left\{
        \begin{array}{ll}
            1 & \text{ if }  i\in e, \\
            0 & \text{ otherwise}
        \end{array}
\right.$,
where $i \in \mathcal{V}$ and $e$ is an hyperedge; \emph{i.e.} a subset of nodes of $\mathcal{V}$.  In our multi-modal setting, we assume a  given  $\mathbf{M}$  modalities. We have for each modality $N$ samples $X = \{x_1,...,x_N\} \in \mathcal{X}$ sampled from a probability distribution $\mathbb{P}$ on $\mathcal{X}$. We then have a multi-modal data collection as $\{X_1,...,X_{s},X_{s+1},..., X_\mathbf{M}\}$. The first $s$ modalities are imaging data (e.g. PET imaging) whilst the remaining are non-imaging data (e.g. genetics).

The first problem we address in our framework is how  to learn embeddings, $\mathbf{v}=f_\theta(x)$, without supervision and invariant to perturbations. That is, given a deep network $f_\theta$, with parameters $\theta$, we seek to map the imaging/non-imaging sample $x$ to a feature space $\mathbf{v}$. As we seek to obtain relevant priors on the data distribution, we also consider a group of $T$ transformations such as there exists
a representation function $\phi : T\times\mathcal{X} \rightarrow \mathcal{X}, (t,x) \mapsto \phi(t,x)$. We then seek to enforce $\phi(t,x) = \phi(x)$ {for all $t\in T$}. We divide our embedding learning problem into two strategies corresponding to each type of data.

For the imaging data, we use  contrastive self-supervised learning (e.g.~\cite{hadsell2006dimensionality,he2020momentum,chen2020simple}) for mapping $X$ to a feature space $\mathbf{v}  = \{\mathbf{v}_1,...,\mathbf{v}_N\}$ with $\mathbf{v}_i=f_{\theta}(x_i)$ such that $\mathbf{v}_i$ better represents $x_i$.
Given $t(x_i)$ and $t'(x_i)$,
where $t, t'\sim T$ are operators that produce two perturbed versions of the given sample (defined following ~\cite{cubuk2020randaugment}), we seek to learn a batch of embeddings that are invariant to any transformation (see Figure 2). Formally, we  compute the following contrastive loss:
\begin{equation}
\begin{aligned} \label{loss_CL}
 \mathcal{L}_{visual}^{(i,j),X_{i\in s}} = - \log \frac{\exp(\mathbf{f}_{i,j} / \tau)}{\sum^{n}_{k=1} \mathbbm{1}_{[k\neq i]}\exp(\mathbf{f}_{i,k} / \tau)}, \text{ where } \mathbf{f}_{i,j} = \frac{\mathbf{v}_i^{\mathrm{T}}\mathbf{v}_j}{||\mathbf{v}_i||\cdot||\mathbf{v}_j||,}
\end{aligned}
\end{equation}

\noindent
where $\tau>0$ is  a temperature hyperparameter and $\mathbf{f}_{i,k}$ follows same cosine similarity definition than $\mathbf{f}_{i,j}$.  We denote the k-nearest neighbors {of an embedding $\mathbf{v}_i$} as $\text{NN}_k{(\mathbf{v}_i)}$. We then construct for each $X_1,...,X_s$ a corresponding  hypergraph $\mathcal{H}_{ij}^{1,..,s}= [\mathbf{v}_i^\top \mathbf{v}_j] \text{ if } \mathbf{v}_i \in \text{NN}_k(\mathbf{v}_j), \text{ otherwise } 0$.

\begin{wrapfigure}{r}{0.40\textwidth} \vspace{-1cm}
  \begin{center}
    \includegraphics[width=0.40\textwidth]{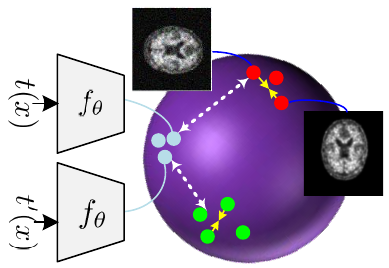}
  \end{center}
  \caption{We seek for distinctiveness. Related samples should have similar features. } \vspace{-0.5cm}
\end{wrapfigure}

For the non-imaging data (e.g. genetics and age), we follow different protocol as perturbing directly the data might neglect the data/structure semantics. We then seek to create a subject-phenotypic relation. To do this, we compute the similarity between the subjects $\mathbf{x}$ and the corresponding phenotypic measures,  to generate the hypergraphs $\mathcal{H}^{s+1,\cdots,\mathbf{M}}$. Given a set of phenotypic measures, we compute a  $\text{NN}_k$ graph    between subjects given the set of measures $\mathbf{z}$
such as $S(\mathbf{\mathbf{x},\mathbf{z}})$ if $\mathbf{x}\in \text{NN}_k(\mathbf{z})$, otherwise 0; being $S$ a similarity function. That is, we enforce the connection of subjects within similar phenotypic measures.
We then seek to perturb  $\mathcal{H}^{s+1,\cdots,\mathbf{M}}$ such that the transformed versions $\hat{\mathcal{H}}^{s+1,\cdots,\mathbf{M}} \sim \mathbbm{q}(\hat{\mathcal{H}}^{s+1,\cdots,\mathbf{M}} | \mathcal{H}^{s+1,\cdots,\mathbf{M}})$ where $\mathbbm{q}(\cdot)$ refers to the transformation drawn from the distribution of the given measure. Our set of transformations are node dropping and edge perturbation. We follow, for the group of transformations $T$, the ratio and dropping probability strategies of \cite{qiu2020gcc,you2020graph}. To maximise the agreement between the computed transformation, we use the same contrastive loss  as defined in \eqref{loss_CL}. We denote the loss as  $\mathcal{L}_{pheno}^{(i,j),\mathcal{H}^{s+1,\mathbf{M}}}$ for  non-imaging data. The final hypergraph, $\mathcal{H}$, is the result of concatenating all hypergraph structures from all modalities, $\mathcal{H}^1,...,\mathcal{H}^{\mathbf{M}}$ given by both imaging and non-imaging data.

\subsection{Dynamically Adjusted Hypergraph  Diffusion Model}
The second key part of our framework is a semi-supervised hypergraph diffusion model. After constructing a robust hypergraph, we now detail how we perform disease classification using only a tiny labelled set.

We consider  a small amount of labelled data $X_L = \{ (x_i ,y_i) \}_{i=1}^{l}$ with provided labels $\mathcal{L} = \{1,..,L\}$  and $y_i \in \mathcal{L}$ along with a large unlabelled set $X_u = \{ x_k \}_{k=l+1}^{m}$. The entire dataset  is then $X= X_L \cup X_U$. The goal is  to use the provided labeled data $X_L$ to infer a function, $f: \mathcal{X} \mapsto \mathcal{Y}$, that maps the unlabelled set to class labels. However,  to obtain such mapping efficiently and with minimum generalisation error is even more challenging when using multi-modal data e.g.~\cite{parisot2018disease}. We cast this problem as a label diffusion process on hypergraphs.

We consider the problem of hypergraph regularisation for semi-supervised classification~\cite{zhou2006learning,hein2013total,feng2019hypergraph}, where we aim to find a function $u^*$ to infer labels for the unlabelled set and enforce smoothness  on the hypergraph. We  seek to solve a problem of the form $u^* = \argmin_{u \in \mathbb{R}^{|\mathcal{V}|} } \{ \mathcal{L}_{emp}(u) + \gamma {\Omega}(u) \}$, where $u$ $\mathcal{L}_{emp}(u)$
refers to an {empirical} loss (e.g., square loss), ${\Omega}(u)$ denotes a regulariser and $\gamma>0$ is a weighting parameter balancing the importance of each term.  Whilst majority of works consider the clique/star expansion~\cite{agarwal2006higher} and follow the principles of~\cite{zhou2006learning}, \textit{we seek to provide a more robust functional} to avoid the inherent disadvantages of such approximations (e.g. bias during the graph diffusion~\cite{hein2013total}).
In particular, we consider the setting of  Rayleigh quotient for $p-$Laplacian. We then seek to estimate, for $p=1$, the particular family of solutions based on the minimisation of the ratio $\Omega(u)= \frac{\text{TV}_H(u)}{\norm{u}}$, where $\text{TV}_H = \sum_{e\in \mathcal{E}}w_e  \text{max}_{i,j \in e}|u_i-u_j|$ is the total variation functional on the hypergraph.
To  do this, we generalise~\cite{feld2019rayleigh} to a hypergraph setting and introduce a dynamic adjustment on the diffusion model, which is based on controlling the predictive uncertainty. Our framework allows for both  binary  and multi-class classification settings. Given a number of epochs $E \in [1, ..., E]$ and following previous notation, we compute an alternating optimisation process as follows.

{\noindent \textbf{Alternating Optimisation for Epoch}$\in [1,..,E]:$}

\textsc{Sub-Problem 1: Diffusion Model Minimisation.} For a given  label, we define a function $u\in\R^n$ over the nodes, and denote its value at node $x$ as $u(x)$. In this binary setting, the objective is to estimate a function $u$ that denotes the presence (resp. absence) of the label for data $x$.  To that end, following~\cite{feld2019rayleigh}, we seek to solve $\min \frac{\text{TV}_H(u)}{\norm{u}}$ with the following semi-explicit flow, in which $u_k$ is the $u$ value at iteration $k$:
\begin{equation}\label{diffusionModel}
\text{Diffusion Model} \left\{\begin{array}{ll} \frac{u_{k+1/2}-u_k}{\delta t}&=\frac{TV_H(u_k)}{\norm{u_k}} (q_k-\tilde q_k)-\subgradp_{k+1/2},\\
u_{k+1}&=\frac{u_{k+1/2}}{\normd{u_{k+1/2}}}
\end{array}\right.
\end{equation}
where   $\subgradp_{k+1/2}\in\partial \text{TV}_H(u_{k+1/2})$,  $q\in\partial \norm{u_k}$ (with $\partial f$ the set of possible subdifferentials of a convex function $f$ defined as $\partial f=\{\subgradp,\, \textrm{ s.t. } \exists u,\, \textrm{ with } \subgradp\in\partial f(u)\}$),  the scaling $d(x)$ is such that $\tilde q_k=\frac{\langle d,q_k\rangle}{\langle d,d\rangle}d$, and $\delta t$ is a a positive time step.  Once the PDE has converged, the output of~\eqref{diffusionModel} is a bivalued function that can be threshold to perform a binary partition of the hypergraph. In order to realise a multi-class labelling, we consider the generalised model of~\cite{aviles2019graphx} that includes as prior the available labels of $X_L$ {in $\mathcal{L}_{emp}(u)$} to guide the partitioning and estimates $\valL$ coupled functions $u^\indexL(x)$  $\indexL=1,\ldots,\valL$. The final labeling is computed as
$\hat{y_i} = \text{argmax}_{i\in \{1,\cdots \valL\}} u^\indexL(x)$ with $L(x)\in \hat{y_i}$.
\textit{Intuition:} Having our constructed hypergraph as in subsection 2.1, we seek to use a tiny labelled set, as we are in a semi-supervised setting, to diffuse the labels for the unlabelled set. A major challenge in semi-supervised learning is how to reduce the prediction bias (e.g.~\cite{arazo2020pseudo}) for the unlabelled set, as it is conditioned solely to the tiny labelled set as prior. We then seek to update the uncertainty of the computed $\hat{y_i}$ via solving Sub-Problem 2.

\textsc{Sub-Problem 2: Uncertainty Hypergraph Minimisation.}
Following the scheme from Figure~\ref{fig::teaser} and notation from previous sections, we first initialise $f_\psi$ using the tiny labelled set available, $X_L$, and then we take $\hat{y_i}$ from Sub-Problem 1 and check the uncertainty of $\hat{y_i}$ by computing the following loss:
\begin{equation}
{ \mathcal{L}_{DYN} (X,Y,\hat{Y};\theta):= \min_{\theta}\sum_{i=1}^{l} \mathcal{L}_{CE}(f_{\theta}(x_i),y_i) +\sum_{i=l+1}^{m} \gamma_i \mathcal{L}_{CE}(f_{\theta}(x_i),\widehat{y}_i)},
\end{equation}
where $\mathcal{L}_{CE}$ is a cross entropy loss and $\gamma$ is the measure of uncertainty via entropy~\cite{kendall2017uncertainties,abdar2021review}
defined as  $\gamma_i = 1 - (H(u_{\hat{y}_i})/\log (L)) $, where $H$ refers to the entropy and  $u_{\hat{y}_i}$ is normalised beforehand.
\textit{Intuition: } From previous step, we obtain an initial prediction for the unlabelled set on the hypergraph. However, in semi-supervised learning due to the tiny labelled set as prior, the unlabelled data tend to have an asymptotic bias in the prediction. We then ensure that there is a high certainty in the unlabelled predictions since early epochs to avoid the propagation of incorrect predictions.

\section{Experimental Results}
In this section, we provide all details of our evaluation protocol.

\smallskip
\smallskip
\textbf{Data Description.} We evaluate our semi-supervised hypergraph framework using the Alzheimer’s disease Neuroimaging Initiative (ADNI) dataset\footnote[1]{Data used were obtained from the Alzheimer’s Disease Neuroimaging Initiative (ADNI) database (adni.loni.usc.edu)}.  ADNI is a multi-centre dataset composed of multi-modal data including imaging and multiple phenotype data. We consider 500 patients using  MRI, PET, demographics and Apolipoprotein E (APOE).  We included APOE as it is known to be a crucial genetic risk factor for developing  Alzheimer’s disease. The dataset contains four categories: early mild cognitive impairment (EMCI), late mild cognitive impairment (LMCI), normal control (NC) and Alzheimer's disease (AD).

\smallskip \noindent
\textbf{Evaluation Protocol.} We design a three part evaluation scheme. Firstly, we follow the majority of techniques convention for binary classification comparing AD vs NC, AD vs EMCI, AD vs LMCI, EMCI vs NC, LMCI vs NC and EMCI vs LMCI (see Supplementary Material for extended results). Secondly, we extended the classification problem  to a multi-class setting including the four classes AD vs NC vs EMCI vs LMCI. We consider this setting, as one of the major challenges in AD diagnosis is to fully automate the task without pre-selecting classes~\cite{goenka2021deep}.

\begingroup
\setlength{\tabcolsep}{2pt}
\renewcommand{\arraystretch}{1}
\begin{table}[t!]
\centering
\resizebox{\columnwidth}{!}{
\begin{tabular}{lcccccccc}
\hline
\multicolumn{1}{c}{\multirow{2}{*}{\textsc{Technique}}} & \multicolumn{3}{c}{\cellcolor[HTML]{EFEFEF}AD vs NC} & \multicolumn{1}{l}{} & \multicolumn{3}{c}{\cellcolor[HTML]{EFEFEF}EMCI vs LMCI} \\ \cline{2-4} \cline{6-8}
\multicolumn{1}{c}{} & ACC & SEN & PPV &  & ACC & SEN & PPV \\ \hline
GNNs~\cite{parisot2018disease}& 81.60$\pm$2.81 & 83.20$\pm$3.10 & 80.62$\pm$2.30 &  & 75.60$\pm$2.50 & 75.20$\pm$3.02 & 75.80$\pm$2.45  \\
HF~\cite{shao2020hypergraph} & 87.20$\pm$2.10 & 88.01$\pm$2.15 & 86.60$\pm$2.60 &  & 80.40$\pm$2.02 & 82.41$\pm$2.14 & 79.23$\pm$2.60  \\
HGSCCA~\cite{shao2021hyper} & 85.60$\pm$2.16 & 87.20$\pm$3.11 & 84.40$\pm$2.15 &  & 76.01$\pm$2.16 & 75.21$\pm$2.01 & 76.42$\pm$2.22 \\
HGNN~\cite{feng2019hypergraph} & 88.01$\pm$2.60 & 90.40$\pm$2.16 & 87.59$\pm$2.42 &  & 80.60$\pm$2.05 & 81.60$\pm$2.54 & 79.60$\pm$2.51 \\
DHGNN~\cite{jiang2019dynamic} & 89.90$\pm$2.40 & 89.60$\pm$2.15 & 90.21$\pm$2.45 &  & 80.80$\pm$2.47 & 82.40$\pm$2.41 & 79.80$\pm$2.76 \\
Ours &  \cellcolor[HTML]{CAFFCA}92.11$\pm$2.03 &  \cellcolor[HTML]{CAFFCA}92.80$\pm$2.16 &  \cellcolor[HTML]{CAFFCA}91.33$\pm$2.43 &  &  \cellcolor[HTML]{CAFFCA}85.22$\pm$2.25 &  \cellcolor[HTML]{CAFFCA}86.40$\pm$2.11 &  \cellcolor[HTML]{CAFFCA}84.02$\pm$2.45 \\ \hline
\end{tabular}}
\caption{Numerical comparison of our technique and existing (graph) and hypergraph techniques. All comparison are run on same conditions. The best results are highlighted in green colour.  } \vspace{-0.7cm}
\label{table:1}
\end{table}
\endgroup

\begingroup
\setlength{\tabcolsep}{4pt}
\renewcommand{\arraystretch}{1.1}
\begin{table}[t!]
\centering
\begin{tabular}{lccc}
\hline
\multicolumn{1}{c}{\multirow{2}{*}{\textsc{Technique}}} & \multicolumn{3}{c}{\cellcolor[HTML]{EFEFEF}LMCI vs NC} \\ \cline{2-4}
\multicolumn{1}{c}{} & ACC & SEN & PPV \\ \hline
GNNs~\cite{parisot2018disease} & 72.40$\pm$2.05 & 70.40$\pm$2.80 & 73.30$\pm$2.04 \\
HF~\cite{shao2020hypergraph} & 77.01$\pm$2.26 & 77.6$\pm$2.15 & 78.22$\pm$2.51 \\
HGSCCA~\cite{shao2021hyper} & 74.00$\pm$2.10 & 74.40$\pm$2.16 & 73.80$\pm$2.17 \\
HGNN~\cite{feng2019hypergraph} & 78.90$\pm$3.01 & 80.01$\pm$2.7 & 78.10$\pm$2.67 \\
DHGNN~\cite{jiang2019dynamic} & 79.20$\pm$2.70 & 80.03$\pm$3.01 & 78.74$\pm$3.22 \\
Ours & \cellcolor[HTML]{CAFFCA}82.01$\pm$2.16 & \cellcolor[HTML]{CAFFCA}84.01$\pm$2.34 & \cellcolor[HTML]{CAFFCA}81.80$\pm$2.55 \\ \hline
\end{tabular}
\caption{Performance comparison of our technique and existing hypergraph models for LMCI vs NC. The results in green colour denotes the highest performance.} \vspace{-0.5cm}
\label{table:2}
\end{table}
\endgroup

To evaluate our model, we performed comparisons with state-of-the-art techniques on hypergraph learning: HF~\cite{shao2020hypergraph}, HGSCCA~\cite{shao2021hyper}, HGNN~\cite{feng2019hypergraph} and DHGNN \cite{jiang2019dynamic}. We also added the comparison against GNNs~\cite{parisot2018disease}, which is based on graph neural networks (GNNs).
We added this comparison as it also considers imaging and non-imaging data. For a fair comparison in performance, we  ran those techniques under same conditions.
The quality check is performed following  standard
\begin{wrapfigure}{r}{0.40\textwidth} \vspace{-0.9cm}
  \begin{center}
    \includegraphics[width=0.40\textwidth]{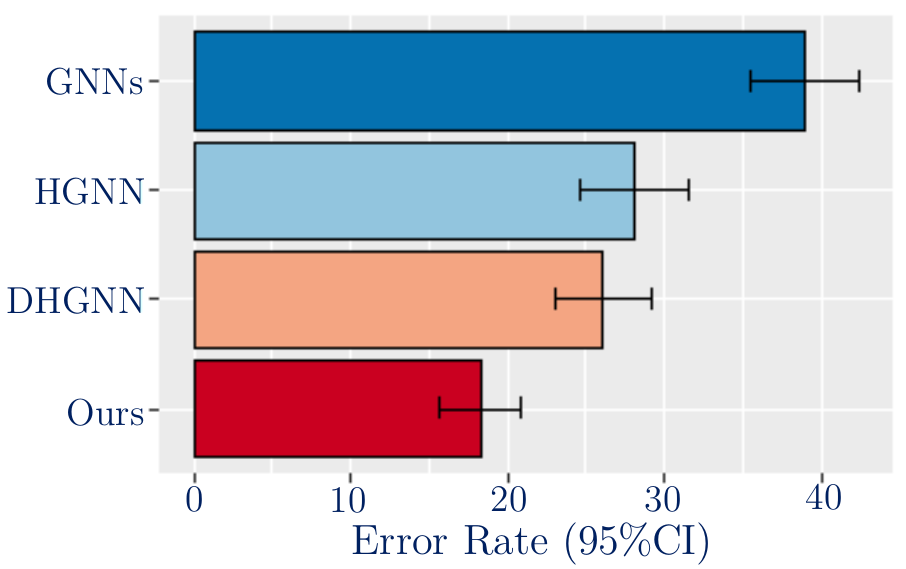}\vspace{-0.4cm}
  \end{center}
  \caption{Performance comparison of ours and SOTA techniques for the four classes case.} \vspace{-0.65cm}
\end{wrapfigure}
convention in the medical domain: accuracy (ACC), sensitivity (SEN), and  positive predictive value (PPV) as a trade-off metric between sensitivity and specificity.
Moreover and guided by the field of estimation statistics, we report along with the error rate the confidence intervals (95\%) when reporting multi-class results.
%
We set the k-NN neighborhood to $k=50$, for the alternating optimisation we used a weight decay of $2\times10^{-4}$ and learning rate was set to 5e-2 decreasing with cosine annealing, and use 180 epochs. For the group of transformations $T$,  we use the strategy of that~\cite{cubuk2020randaugment} for the imaging data whilst for the non-imaging data the ratio and dropping probability strategies of \cite{qiu2020gcc,you2020graph}.
Following standard protocol in semi-supervised learning, we randomly
select the labelled samples over five repeated time and then report the mean of the metrics along with the standard deviation.

\smallskip \noindent
\textbf{Results \& Discussion.}
We began by evaluating our framework following the binary comparison of AD vs NC, EMCI vs LMCI and LMCI vs NC (see supplementary material for extended results). We report a detailed quantitave analysis to understand the performance of ours and compared techniques. We use 15\% of labelled data (see supplementary material).   The results are reported in Tables~\ref{table:1} \&~\ref{table:2}. In a closer look at the results, we observe that our technique reports the best performance for all comparison and in all metrics. We also observe that the GNNs~\cite{parisot2018disease} reported the worst performance. This is due to the fact that GNNs does not allow for higher-order relations on the graph.  We highlight that whilst the other techniques reported a good performance, our technique substantially improved over all methods (with statistical significance, see supplementary material). We underline that all compared techniques follows the approximation of~\cite{zhou2006learning}, which introduces bias during the hypergraph partition. We showed that our diffusion model, that follows different principles, mitigates that problem.

\begingroup
\setlength{\tabcolsep}{2pt}
\renewcommand{\arraystretch}{1.1}
\begin{table}[t!]
\centering
\begin{tabular}{lccc}
\cline{1-2} \cline{4-4}
\multicolumn{1}{c}{} & \cellcolor[HTML]{EFEFEF}AD vs NC   vs MCI &  & \multicolumn{1}{c|}{\cellcolor[HTML]{EFEFEF}AD vs NC   vs EMCI vs LMCI} \\ \cline{2-2} \cline{4-4}
\multicolumn{1}{c}{\multirow{-2}{*}{\textsc{Technique}}} & \textsc{Error Rate \& 95\%CI} &  & \textsc{Error Rate \& 95\%CI} \\ \cline{1-2} \cline{4-4}
GNNs~\cite{parisot2018disease} & 36.19$\pm$4.45 &  & 39.01$\pm$3.12 \\
HGNN~\cite{feng2019hypergraph} & 26.35$\pm$3.20 &  & 28.09$\pm$3.65 \\
DHGNN~\cite{jiang2019dynamic} & 23.10$\pm$2.60 &  & 26.25$\pm$2.55 \\
Ours &  \cellcolor[HTML]{CAFFCA}16.25$\pm$2.22 &  &  \cellcolor[HTML]{CAFFCA}18.31$\pm$2.45 \\ \cline{1-2} \cline{4-4}
\end{tabular}
\caption{Error rate comparison of our technique against existing models. The results in green denotes the best performance.} \vspace{-0.2cm}
\label{table:3}
\end{table}
\endgroup

\begin{figure}[t!]
\centering
\includegraphics[width=1\textwidth]{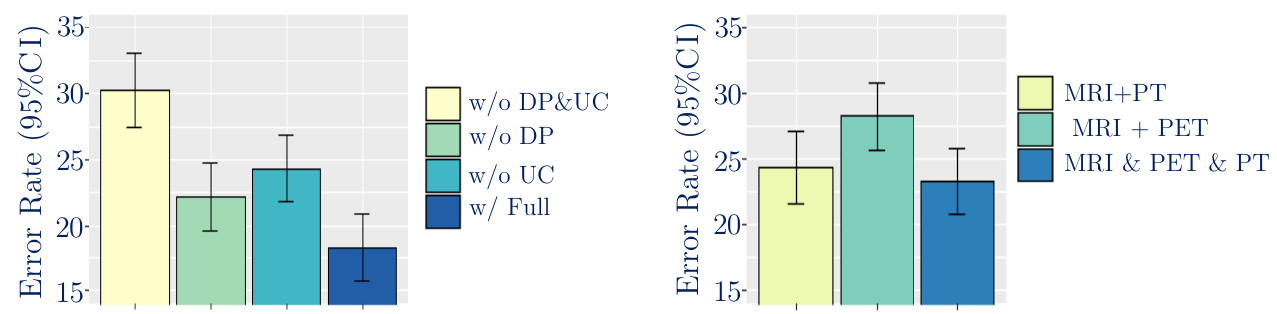}
\caption{(left side) Ablation study for the components of our technique. (right side) Performance comparison using different type of multi-modal data.} \vspace{-0.5cm}
\label{fig::ablation2}
\end{figure}

To further support our previous results, we also report results in the more challenging setting of multi-class classification for AD diagnosis. The results are reported in Table~\ref{table:3}. We observe that our technique performance is consistent with previous results. That is, our technique reported the lowest error rate for both multi-class cases. Our technique decreased the error rate for more than 40\% for all techniques. Figure 3 complements these results by reporting the performance with respect to the four classes case.

Finally, to support the design of our technique, we performed two ablations studies regarding our design and the modalities used. We start by evaluating the influence of our full model against  without our dual perturbed embedding strategy (displayed as DP) and our uncertainty (denoted as UC)
scheme. The results are reported at the left side of Figure~\ref{fig::ablation2} on the four classes case.
Whilst our dual strategy substantially improves the performance, we observe that our uncertainty scheme has a major repercussion. The intuition is that our diffusion model ensures, at early epochs, that there is a high certainty in the prediction avoiding incorrect prediction in subsequent epochs. We also include an ablation study regarding the impact of the modalities (PT refers to phenotypic data). From  the right side of Figure~\ref{fig::ablation2}, we observe that including phenotypic data has a greater positive impact on the performance than using only imaging data.

\section{Conclusion}
We proposed a novel semi-supervised hypergraph framework for Alzheimer’s disease diagnosis. Unlike existing techniques, we introduce a dual embedding strategy. As phenotypic data highly differs from imaging data. Moreover, in contrast to existing techniques that follow the hypergraph approximation of~\cite{zhou2006learning}. We introduce a better diffusion model, which solution is provided by a semi-explicit flow in the fra hypergraph learning. From our results, we showed that our technique outperforms other hypergraph techniques.  Future work includes a more extensive clinical evaluation with public and in-home datasets.

\subsubsection{Acknowledgements}
\sloppy
AIAR  acknowledges support from CMIH and
CCIMI, University of Cambridge. CR acknowledges support from the CCIMI and the EPSRC grant  EP/W524141/1 ref. 2602161.
ZK acknowledges support from the BBSRC (H012508, BB/P021255/1), Wellcome Trust (205067/Z/16/Z, 221633/Z/20/Z) and Royal Society (INF/R2/202107).
CBS acknowledges the Philip Leverhulme Prize, the EPSRC  fellowship EP/V029428/1, EPSRC grants EP/T003553/1, EP/N014588/1, Wellcome Trust 215733/Z/19/Z and 221633/Z/20/Z, Horizon 2020  No. 777826 NoMADS and the CCIMI.

\bibliographystyle{splncs04}
\bibliography{paper2386_bib}

\clearpage
\newpage
%
%
%
%

{ {\section*{\Large \centering Supplemental Material \\ \centering Multi-Modal Hypergraph Diffusion Network with Dual Prior for Alzheimer Classification }}}

\vspace{0.3cm}
\begin{center}
{ \centering Angelica I. Aviles-Rivero$^1$, Christina Runkel$^1$, Nicolas Papadakis$^2$} \\
 {\centering Zoe Kourtzi$^3$   and Carola-Bibiane Schonlieb$^1$ }
\end{center}

\vspace{0.3cm}
{
\centering $^1$DAMTP, University of Cambridge, UK  \email{\{ai323,cr661,cbs31\}@cam.ac.uk}\\
$^2$IMB, Universite de Bordeaux, France \email{nicolas.papadakis@math.u-bordeaux.fr} \\
\centering $^3$Department of Psychology, University of Cambridge, UK \email{zk240@cam.ac.uk}
}

%
%
%


\vspace{0.5cm}
\section*{Supplementary Numerical Results \& Details}
\begin{figure}[h]
\centering
\includegraphics[width=0.9\textwidth]{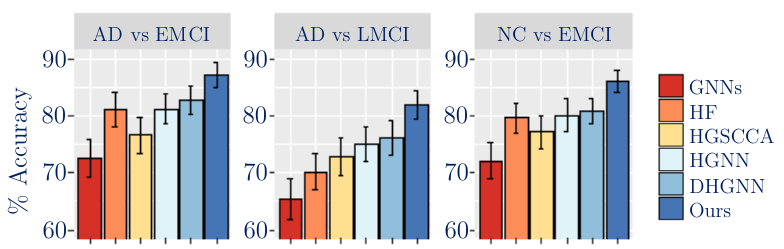}
\caption{Performance comparison of our technique vs existing techniques. All results are run under the same conditions. }
\label{fig::suppl1}
\end{figure}

\textbf{A. Supplementary Results.} In this section, we extend the numerical results of Table 1 from the main paper.  Our results report the performance of AD vs EMCI, AD vs LMCI and NC vs EMCI. The comparison includes GNNs~\cite{parisot2018disease},HF~\cite{shao2020hypergraph}, HGSCCA~\cite{shao2021hyper}, HGNN~\cite{feng2019hypergraph} and DHGNN~\cite{jiang2019dynamic}.
Following standard protocol in semi-supervised learning, we report the results over five repeated randomly selected labelled sets, and report the mean of metrics along with the standard deviation. The results are reported in Fig.~\ref{fig::suppl1}. In a closer look at the results, we observe that the performance behaviour of our technique is in line to the reported in the main paper.
That is, our technique consistently outperforms by a significant margin the compared techniques up to $\sim19\%$ with respect to the GNNs and in the range of $\sim8$  to $\sim16\%$ compared with other hypergraph techniques.

\medskip
\textbf{B. Statistical Tests.} To further support our technique, we performed a set of non-parametric statistical tests.
For the binary comparisons, we performed a Wilcoxon test, and we found that the difference in performance of our technique vs the compared ones is statistically significant ($p<0.0001$).
For the multi-class setting (4-classes case), we use the non-parametric Friedman test, followed by a pair-wise comparison with adjusted $p$-values using    the Bonferroni method.
\begin{wrapfigure}{r}{0.40\textwidth} \vspace{-0.6cm}
  \begin{center}
    \includegraphics[width=0.40\textwidth]{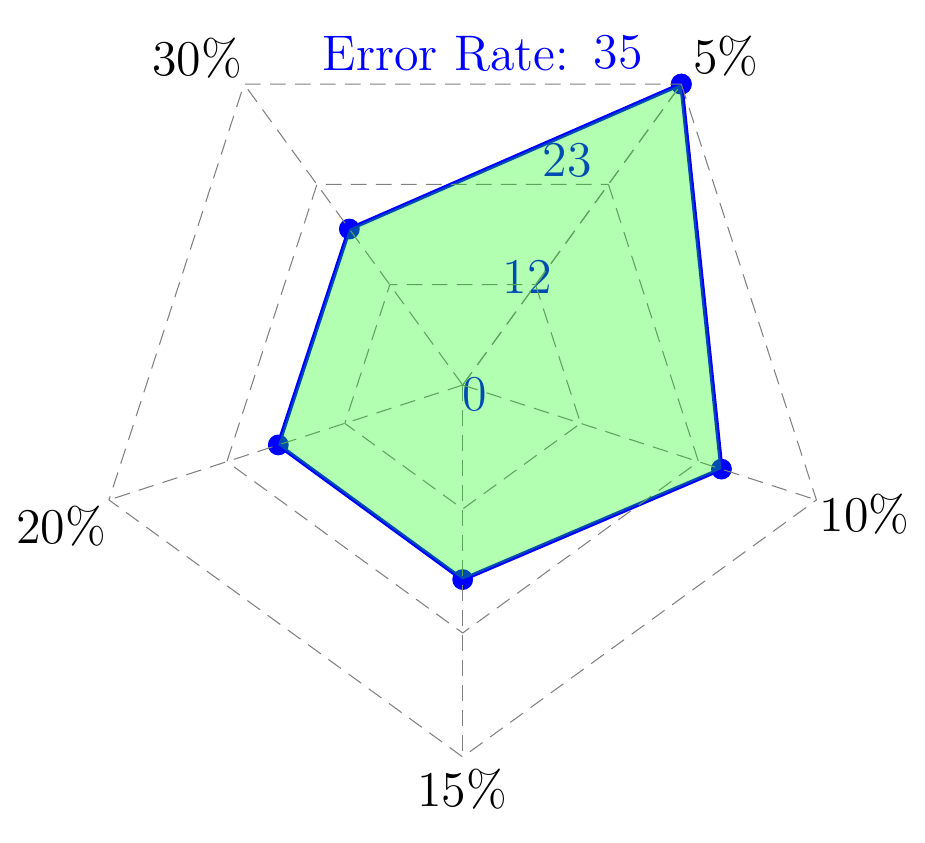}
  \end{center}
  \caption{Error rate vs different label counts. }
  \vspace{-0.8cm}\label{fig:suppl2}
\end{wrapfigure}
We found that our technique was statistically significant, ($\chi^2=18.1, p=0.0002$), different with respect to all compared techniques.
Hence, the generalisation and robustness of our technique is demonstrated.

\medskip
\textbf{C. Supplementary Details.} We used ADNI-2 in our experiments as it provides an explicit separation between EMCI and LMCI. Moreover, for the imaging data we used Florbetapir (FBP) PET and T1 MR images.  Our experiments from the main paper were computed using only 15\% of labelled data. However, we provide an ablation study regarding the behaviour of our technique under different labelled counts. The results are reported 
for the four classes case, and in terms of error rate. The results are displayed in Fig.~\ref{fig:suppl2}. We use 15\% of labelled data for the following reasons. Firstly, we want to use a labelled rate as lower as possible. Secondly, as displayed in our results 15\% is a good trade-off between performance and  label rate. We can observe that labelled rates greater than 25\% only provides a tiny improvement in performance.  This performance effect is well-known in semi-supervised learning, as the transductive effect e.g.~ \cite{joachims1999transductive},  and also observed in our experiments.

\end{document}